\documentclass[letterpaper, 10 pt, conference]{ieeeconf}
\IEEEoverridecommandlockouts                              
\usepackage{amsmath}
\usepackage{graphicx}
\usepackage{caption}
\DeclareCaptionFont{mysize}{\fontsize{8}{9.6}\selectfont}
\usepackage{amssymb}
\usepackage{float}
\usepackage{mathtools}
\usepackage{amsmath}
\usepackage{anyfontsize}
\usepackage{xcolor}
\usepackage{colortbl}
\usepackage{lipsum}
\usepackage{fontenc}
\usepackage{hyphenat}
\usepackage{mathrsfs} 
\usepackage{enumerate}
\usepackage{subfig}
\usepackage[abs]{overpic}
\usepackage[utf8]{inputenc}
\usepackage{dtk-logos} 
\usepackage{lipsum}
\usepackage{mathtools, nccmath}
\usepackage{cuted}
\usepackage{authblk}
\input{insbox.tex}

\usepackage{algorithm}
\usepackage{algorithmic}
\makeatletter
\newcommand\fs@norules{\def\@fs@cfont{\bfseries}\let\@fs@capt\floatc@ruled
  \def\@fs@pre{}%
  \def\@fs@post{}%
  \def\@fs@mid{\kern3pt}%
  \let\@fs@iftopcapt\iftrue}
\makeatother
\floatstyle{norules}
\restylefloat{algorithm}

\newcommand{\bTheta}{\boldsymbol{\Theta}}

\usepackage{epstopdf}
\usepackage{listings}
\usepackage{xcolor}
\usepackage{matlab-prettifier}
\usepackage{color} 
\definecolor{mygreen}{RGB}{28,172,0} 
\definecolor{mylilas}{RGB}{170,55,241}
\title{\huge \bf
Learning Discrepancy Models From Experimental Data 
}

\author{Kadierdan Kaheman$^{1}$, Eurika Kaiser$^{1}$ , Benjamin Strom$^{1}$, J. Nathan Kutz$^{2}$  and Steven L. Brunton$^{1}$ 
	\thanks{$^{1}$ Kadierdan Kaheman, Eurika Kaiser, Benjamin Strom, and Steven L. Brunton are with the Department of Mechanical Engineering, University of Washington, Seattle,
		WA, 98105, USA.}%
	\thanks{$^{2}$ J. Nathan Kutz is with the Department of Applied Mathematics, University of Washington, Seattle,
		WA, 98105, USA. }
	\thanks{E-mails: \{kadierk, eurika, strombw, kutz, sbrunton\}@uw.edu}%
}

\begin{document}

\maketitle
\vspace{-1ex}
\begin{abstract}
First principles modeling of physical systems has led to significant technological advances across all branches of science. 
For nonlinear systems, however,  small modeling errors can lead to significant deviations from the true, measured behavior. 
Even in mechanical systems, where the equations are assumed to be well-known, there are often model discrepancies corresponding to nonlinear friction, wind resistance, etc. 
Discovering models for these discrepancies remains an open challenge for many complex systems. 

In this work, we use the sparse identification of nonlinear dynamics (SINDy) algorithm to discover a model for the discrepancy between a simplified model and measurement data.  
In particular, we assume that the model mismatch can be sparsely represented in a  library of candidate model terms. 
We demonstrate the efficacy of our approach on several examples including experimental data from a double pendulum on a cart.
We further design and implement a feed-forward controller in simulations, showing improvement with a discrepancy model. 
\end{abstract}

\section{Introduction}
\label{sec1}

There are currently two dominant approaches in dynamical systems modeling: 1) deriving equations of motion from governing physical principles, and 2) data-driven system identification. 
These two approaches are typically used in isolation to develop an end-to-end model. 
Data-driven models, based on emerging techniques in machine learning, are promising for the description of complex systems~\cite{Brunton2019book}.  
However, these models are often opaque, and it is difficult to capture the correct dynamical system structure to satisfy basic constraints and conservation laws. 
In contrast, first-principles physics models capture constraints and conservation laws by design~\cite{Marsden:MS}, but for complex systems, these models are either overly simplistic or exceedingly expensive to resolve all multiscale interactions. 
In this work, we propose a hybrid modeling approach, where we employ data-driven techniques to model the discrepancy between a simplified or insufficient physical model and observed measurements.  
We employ the sparse identification of nonlinear dynamics (SINDy)  framework~\cite{SINDY} to discover parsimonious and interpretable discrepancy models.
Thus, we leverage prior knowledge of the simplified physics, while more accurately modeling details of the true system.  

As an illustrative example for this paper, we consider a double pendulum on a cart experiment.  
We typically model this type of experiment as a simple mechanical system with a few degrees of freedom, described by either a Hamiltonian or Lagrangian~\cite{Marsden:MS}. 
However, this model neglects real-world effects such as nonlinear friction, bearing chatter, and wind resistance.  
These factors may all reduce control performance, for example, in model predictive control (MPC)~\cite{camacho2013model,eren2017jgcd,Kaiser2018prsa}, where prediction errors adversely affect robustness~\cite{MethodologyMPCMismatch}. 
For this specific system, designing a feed-forward control law to swing up the pendulum from rest requires a highly accurate model. 
Even a small mismatch in the system model may result in a considerable deviation in the computed trajectory, since it is a chaotic system. 
It is also challenging to obtain a data-driven model of this system with the correct structure, so  it is beneficial to incorporate prior physical knowledge in the form of a simplified Hamiltonian or Lagrangian.   

\subsection{Problem statement: Discrepancy modeling}
There are several reasons why model discrepancies occur~\cite{Bayesiancalibrationofcomputermodels,QuanModelUncertainty}. 
First, there may be measurement noise and exogenous disturbances.  
In this case, the Kalman filter may be thought of as a discrepancy model where the mismatch between a simplified model and observations is assumed to be a Gaussian process~\cite{Kalman1960jfe}.  
Next, the parameters of the system may be inaccurately modeled. 
Even worse, the structure of the model may not be correct, either because important terms are missing or erroneous terms are present.  
This is known as \emph{model inadequacy} or model structure mismatch.  
Other challenges include incomplete measurements and latent variables, delays, and sensitive dependence on initial data in chaotic systems. 

In this work, we focus on parameter and structural uncertainties, although a broader framework is the subject of ongoing work. 
We consider dynamical systems of the form
\begin{align}
    \frac{d}{dt}\boldsymbol{x} = \boldsymbol{f}(\boldsymbol{x},{\bf u};\boldsymbol{\mu}),
\end{align}
where $\boldsymbol{x}\in\mathbb{R}^n$ is the state, $\boldsymbol{u}\in\mathbb{R}^r$ is the control input, $\boldsymbol{\mu}\in\mathbb{R}^p$ are the parameters, and $\boldsymbol{f}$ are the dynamics. 
We assume full-state measurements, although generally the state must be estimated from  limited measurements. 

The discrepancy modeling problem seeks to model the difference between a quantity of interest $\boldsymbol{\phi}(t)$ from a physical model $\boldsymbol{\phi}_m(t)$ and the observed value $\boldsymbol{\phi}_o(t)$:
\begin{align}
    \delta\boldsymbol{\phi}(t) = \boldsymbol{\phi}_o(t) - \boldsymbol{\phi}_m(t),
\end{align}
where $\delta\boldsymbol{\phi}$ is the discrepancy. 
Here, we consider the quantity of interest $\boldsymbol{\phi}(t)$ to be the dynamics themselves, or more precisely the rate of change of the state in time.  However, this framework is more general and can incorporate several other forms of model discrepancy. For example, $\boldsymbol{\phi}(t)$ may be a conserved quantity, such as the Hamiltonian, from which the dynamics are derived.

\subsection{Related work}
Accurate system modeling is a critical task in science and engineering, including for autonomous robotics~\cite{RobotMPC1,RobotMPC2} and process control~\cite{Chemical1,Chemical2,HandlingModelPlantMismatchinStateEstimation}. 
A model that is able to accurately predict the future state of a system is imperative for prediction and control. 
However, the high-dimensional, nonlinear, and multi-scale nature of many systems renders modeling a challenging task. 

System identification has reached a high degree of maturity encompassing myriad techniques to identify linear and nonlinear systems~\cite{Nelles2013book,ljung2010arc} from data, including state-space modeling via the eigensystem realization algorithm (ERA)~\cite{ERA:1985} and other subspace identification methods, Volterra
series~\cite{Brockett1976automatica,maner1994nonlinear}, linear and nonlinear autoregressive models~\cite{Akaike1969annals} (e.g., ARX, ARMA, NARX, and NARMAX), and neural network models~\cite{lippmann1987introduction,draeger1995model}, to name only a few. Machine learning techniques such as manifold learning and non-parametric modeling have also been useful for identifying nonlinear systems~\cite{principe1998ieee,ko2007ieee}. 

There is an increasing shift from black-box modeling to developing models that are physically intuitive and interpretable, as well as models that are constrained models with known prior information.
For instance, genetic programming can be used to infer governing equations from data~\cite{Bongard2007pnas,Schmidt2009science}.  
The recent SINDy approach~\cite{SINDY} identifies parsimonious and interpretable models by promoting sparsity, and it has been extended to incorporate the effect of control~\cite{Kaiser2018prsa} and to take into account expert knowledge, such as symmetries and conservation laws~\cite{Loiseau2017jfm}.  
However, it may be challenging to identify certain types of systems, such as those containing rational function nonlinearities~\cite{Mangan2016ieee}. 
Instead of modeling the system entirely from data, when partial information is available, such as an idealized Hamiltonian, data-driven modeling procedures may focus only on modeling the discrepancy. 

One way to compensate for model discrepancy is with Bayesian approaches~\cite{Bayesiancalibrationofcomputermodels}, where a model discrepancy function is learned and the model output uncertainty is quantified from data~\cite{ParameterEstimationforPredictiveSimulation}. However, this method requires the selection of a prior form for the model discrepancy function, which is difficult due to the lack of physical knowledge about the model structure~\cite{SelectionofmodeldiscrepancypriorsinBayesiancalibration}. Furthermore, this method may potentially introduce bias in the model parameters if calibration is performed without knowledge of the model discrepancy~\cite{Learningaboutphysicalparameters,ParameterEstimationandModelDiscrepancyinControlSystems}. Alternatively, reinforcement learning can be used to learn the mismatch between a model and the actual dynamics, for example in robotics~\cite{Model-PlantMismatchCompensationUsingReinforcementLearning}. The learned dynamics are then used with the conceptual model to control the real system. However, neither Bayesian methods nor reinforcement learning can provide an interpretable representation for the model mismatch, and thus conceal the physical meaning of the discrepancy model discovered.

\vspace{-1ex}
\subsection{Contributions of this work}
\vspace{-0.5ex}
In the present work, we leverage SINDy to compensate for model parameter and structure mismatch given an imperfect model of the system. 
This serves several purposes: (1) Prior partial knowledge on the system or a previously learned model may be available and can be incorporated to aid the modeling process and improve prediction accuracy. (2) The SINDy algorithm suffers from the curse of dimensionality due to the growth of the library size with increasing number of variables in the system, which makes it challenging to discover the full governing equations when a large library size is required. However, learning the model mismatch significantly reduces this burden, focusing SINDy only on modeling the mismatch. 
(3) Learning interpretable representations of the model mismatch may inform physical intuition and generalize beyond the training data. 
We will demonstrate that SINDy is able to model discrepancies such as incorrect system parameters and model inadequacy (or structure) errors. 
The learned SINDy discrepancy model can then be used to enhance the imperfect model, providing an improved description of the system dynamics.

\section{Background: Sparse Identification of Nonlinear Dynamics}
\label{sec2}

Here, we briefly introduce the SINDy algorithm for identifying dynamics from data~\cite{SINDY}.
We consider a system of coupled nonlinear ordinary differential equations (ODE):
\begin{equation}
    \frac{d}{dt}{\boldsymbol{x}}(t)=\boldsymbol{f}(\boldsymbol{x}(t);\boldsymbol{\mu}),
    \label{eq4}
\end{equation}
with time-dependent state  ${\boldsymbol{x}(t)= [{x_{1}(t)},{\ldots}, {x_{n}(t)}]^{T} \in \mathbb{R}^{n}}$. The function $\boldsymbol{f}(\boldsymbol{x}(t);\boldsymbol{\mu})$ describes the underlying dynamics of the system; it is possible to extend this formulation to include control~\cite{Kaiser2018prsa}, although we omit it here for simplicity. 
$\boldsymbol{f}$ is often sparse in the space of candidate functions. 

If this assumption holds, sparse regression~\cite{SINDY,Zheng2019ieeeacess} can be used to infer $\boldsymbol{f}$ from measurement data as explained in the following.

\begin{enumerate}[Step 1:]
\item Form a data matrix $\boldsymbol{X}\in \mathbb{R}^{m\times n}$ and matrix of derivatives $\dot{\boldsymbol{X}}\in \mathbb{R}^{m\times n}$ sampled at times $t_1,\ldots,t_m$:
\begin{subequations}
\begin{align}
\boldsymbol{X}&=[\boldsymbol{x}(t_1)\ \boldsymbol{x}(t_2)\, \cdots\, \boldsymbol{x}(t_m)]^T,\\
\dot{\boldsymbol{{X}}}&=[\dot{\boldsymbol{x}}(t_1)\ \dot{\boldsymbol{x}}(t_2)\, \cdots\, \dot{\boldsymbol{x}}(t_m)]^T.
\end{align}
\label{eq5a6}
\end{subequations}

\item Construct a library of candidate functions $\boldsymbol{\Theta}(\boldsymbol{X})$
\begin{equation}
    \boldsymbol{\Theta}(\boldsymbol{X})=[{\theta}_1(\boldsymbol{X})\ {\theta}_2(\boldsymbol{X})\ {\theta}_3(\boldsymbol{X})\, \cdots\, {\theta}_p(\boldsymbol{X})],
    \label{eq7}
\end{equation}
where ${\theta}_i(\boldsymbol{x})$ is a candidate function that may be present in the dynamics $\boldsymbol{f}$.  For example, trigonometric functions  ${\theta}_i(\boldsymbol{x})=\sin (x_j)$ or polynomial functions ${\theta}_i(\boldsymbol{x})=x_jx_k^2$ may be considered.  The data matrix $\boldsymbol{\Theta}$ is obtained by evaluating these functions at all $m$ entries of the data matrix $\boldsymbol{X}$.

\item Form the sparse regression problem such that
\begin{equation}
    \dot{\boldsymbol{X}}=\boldsymbol{\Theta}(\boldsymbol{X}) \boldsymbol{\Xi}, 
    \label{eq8}
\end{equation}
where the matrix $\boldsymbol{\Xi}=[{\boldsymbol{\xi}_{1}},\ldots,{\boldsymbol{\xi}_{n}}]\in \mathbb{R}^{p\times n}$ comprises sparse vectors ${\boldsymbol{\xi}_{i}}\in \mathbb{R}^{p\times 1}$ that select the active terms in the library $\boldsymbol{\Theta}$. These are found via sparse optimization~\cite{Zheng2019ieeeacess} to minimize the following:
 \begin{equation}
     \boldsymbol{\xi}_i=\underset{\boldsymbol{\xi}_i^{\prime}}{\operatorname{argmin}}\left\|\boldsymbol{\Theta}(\boldsymbol{X}) \boldsymbol{\xi}_i-\dot{\boldsymbol{X}}\right\|_{2}+\lambda\left\|\boldsymbol{\xi}_i\right\|_{1}.
     \label{eq3}
 \end{equation}

\end{enumerate}

\section{\hspace{-0.1cm}Data-Driven Discovery of Model Discrepancy}
\label{sec3}
In this section, we use SINDy to model discrepancies. 
\vspace{-1ex}
\subsection{Discrepancy modeling for systems without control}
Although it is possible to model any physical quantity, we consider modeling the dynamics themselves~\cite{SINDY}.  Consider noisy measurements from a true dynamical system $\boldsymbol{f}$
\begin{equation}
    \boldsymbol{\phi}_o(t)=\boldsymbol{f}(\boldsymbol{x}(t);\boldsymbol{\mu})+\epsilon,
    \label{eq9}
\end{equation}
where $\epsilon\in \mathbb{R}^n$ is the measurement noise. The model output for this system is represented as
\begin{equation}
    \boldsymbol{\phi}_m(t)=\boldsymbol{f}_m(\boldsymbol{x}(t);{\boldsymbol{\mu}}_1).
    \label{eq10}
\end{equation}

Parameter error,  $\boldsymbol{\mu}\neq{\boldsymbol{\mu}}_1$, model inadequacy, $\boldsymbol{f}{\not =}\boldsymbol{f}_m$, and measurement error will cause a model mismatch, given by
\begin{equation}
    \delta{\boldsymbol{\phi}}(t)=\boldsymbol{\phi}_o(t)-\boldsymbol{{\phi}}_m(t) 
    = \boldsymbol{f}(\boldsymbol{x}(t);\boldsymbol{\mu}) -
    \boldsymbol{f}_m(\boldsymbol{x}(t);{\boldsymbol{\mu}}_1).
    \label{eq11}
\end{equation}

We then construct a model for the discrepancy $\delta{\boldsymbol{\phi}}(t)$ as a function of the state and new parameters ${\boldsymbol{\mu}}_2$, capturing the parameter and structure mismatch:
\begin{equation}
    \delta\boldsymbol{\phi}(t)=\boldsymbol{g}(\boldsymbol{x}(t);{\boldsymbol{\mu}}_2).
    \label{eq12}
\end{equation}
Model error data is collected in
\begin{equation}
     \label{eq16}
     \delta\boldsymbol{\boldsymbol{\Phi}}=[\delta\boldsymbol{\phi}(t_1)\ \delta\boldsymbol{\phi}(t_2) \,\cdots\ \delta\boldsymbol{\phi}(t_m)]^T,
\end{equation}
where $\delta\boldsymbol{\boldsymbol{\Phi}}\in \mathbb{R}^{m\times n}$. 

Given data $\delta\boldsymbol{\Phi}$ and $\boldsymbol{X}$, we use  SINDy  to sparsely represent $\boldsymbol{g}(\boldsymbol{x}(t);\boldsymbol{\mu}_2)$ in a library of candidate functions $\boldsymbol{\Theta}(\boldsymbol{\boldsymbol{X}})$. 
Constructing this library typically requires some prior knowledge of the system to select a suitable basis in which the discrepancy model will be sparse.
The sparse regression problem is then formulated as 
\begin{equation}
    \delta\boldsymbol{\boldsymbol{\Phi}}=\boldsymbol{\Theta}(\boldsymbol{\boldsymbol{X}})\boldsymbol{\Xi}.
    \label{eq15}
\end{equation}
\begin{figure*}
	\vspace{1ex}
    \centering
    \includegraphics[width=0.995\textwidth, trim=0 0 0 0cm, clip=truerue]{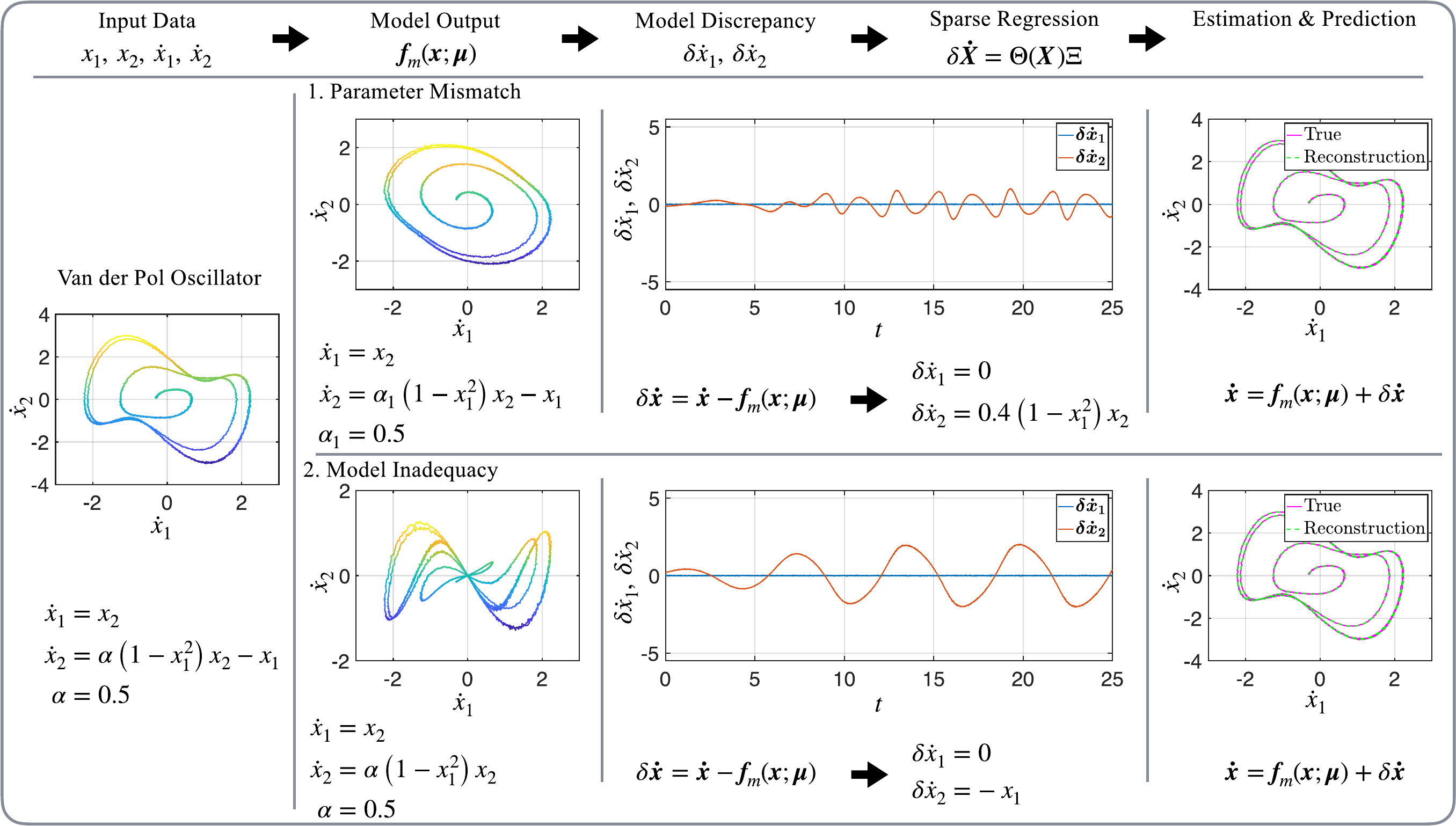}
    \vspace{-3ex}
    \caption{Illustration of SINDy to discover the system-model mismatch for the Van der Pol oscillator. The system is simulated to generate the measurement data. Measurement data $x_1$ and $x_2$ are provided to calculate the output estimated by our imperfect model. Sparse regression is used to infer the discrepancy model for the difference between the actual output and estimated output. The discrepancy model is then combined with the imperfect model to provide a better estimation of system dynamics. The model is then cross-validated on a new initial \textcolor{red}{condition } $\boldsymbol{x}_0=(-0.2,-0.3)$. As we can see, the model discovered by SINDy is able to compensate for the discrepancy between the actual system and flawed model.}
    \label{fig2}
    \vspace{-3ex}
\end{figure*}
Realistically, we will only have access to measurements of $\boldsymbol{x}(t)$, from which we may derive $\frac{d}{dt}\boldsymbol{x}(t)$, and the quantity of interest becomes $\boldsymbol{\phi}(t) \approx \frac{d}{dt}{\boldsymbol{x}}(t)$. The model output is $\boldsymbol{\phi}_m=\boldsymbol{f}_m(\boldsymbol{x};{\boldsymbol{\mu}}_1)$, the discrepancy is $\delta \boldsymbol{\phi}=\frac{d}{dt}\boldsymbol{x}(t)-\boldsymbol{f}_m(\boldsymbol{x};{\boldsymbol{\mu}}_1)$, and we have 
\begin{equation}
    \delta\dot{\boldsymbol{X}}:=\dot{\boldsymbol{X}} - \boldsymbol{f}_m(\boldsymbol{X};{\boldsymbol{\mu}}_1)
    =\boldsymbol{\Theta}(\boldsymbol{X}) \boldsymbol{\Xi}.
    \label{eq18}
\end{equation}
We solve for the sparsest coefficient matrix $\boldsymbol{\Xi}$ that satisfies Eq.~\eqref{eq18} using the SINDy approach.  
 A schematic of the method is displayed in Fig.~\ref{fig2}.

\vspace{-1ex}
\subsection{Discrepancy modeling for systems with control}
\label{SecSINDyc}
We now extend the formulation above to identify dynamics that are affected by a control input $\boldsymbol{u}(t)\in \boldsymbol{\mathbb{R}}^q$:
\begin{equation}
    \boldsymbol{\phi}_o(t)=\boldsymbol{f}(\boldsymbol{x}(t),\boldsymbol{u}(t);\boldsymbol{\mu})+\epsilon,
    \label{sceq1}
\end{equation}
The model output for this system is represented as
\begin{equation}
    \boldsymbol{\phi}_m(t)=\boldsymbol{f}_m(\boldsymbol{x}(t),\boldsymbol{u}(t);{\boldsymbol{\mu}}_1).
    \label{eq2}
\end{equation}
As with the uncontrolled case, due to the model inadequacy or parameter error, there will be a discrepancy
\begin{equation}
    \label{sceq3}
    \begin{split}
        \delta{\boldsymbol{\phi}}(t)&=\boldsymbol{\phi}_o(t)-\boldsymbol{{\phi}}_m(t)\\ 
    &= \boldsymbol{f}(\boldsymbol{x}(t),\boldsymbol{u}(t);\boldsymbol{\mu}) -
    \boldsymbol{f}_m(\boldsymbol{x}(t),\boldsymbol{u}(t);{\boldsymbol{\mu}}_1).
    \end{split}
\end{equation}
Our goal is \textcolor{red}{to} model $\delta{\boldsymbol{\phi}}(t)$ and we further assume it is a function of the system state $\boldsymbol{x}$, control input $\boldsymbol{u}$, and new parameters $\boldsymbol{\mu}_2$: 

\begin{equation}
    \label{sceq4}
    \delta{\boldsymbol{\phi}}(t)=\boldsymbol{g}(\boldsymbol{x}(t),\boldsymbol{u}(t);{\boldsymbol{\mu}}_2).
\end{equation}
To achieve this goal, we form a sparse regression problem
\begin{equation}
    \label{sceq5}
    \delta \boldsymbol{\Phi}=\bTheta(\boldsymbol{X},\boldsymbol{U})\boldsymbol{\xi},
\end{equation}
where  $\boldsymbol{U}\in \mathbb{R}^{m\times r}$ is
\begin{equation}
    \label{sceq6}
    \boldsymbol{U}=[\boldsymbol{u}(t_1)\ \boldsymbol{u}(t_2) \cdots\ \boldsymbol{u}(t_m)]^T.
\end{equation}
As before, we will actually be observing $\boldsymbol{\phi}(t) \approx \frac{d}{dt}{\boldsymbol{x}}(t)$ with model $\boldsymbol{\phi}_m=\boldsymbol{f}_m(\boldsymbol{x},\boldsymbol{u};{\boldsymbol{\mu}}_1)$.  The discrepancy becomes $\delta \boldsymbol{\phi}=\frac{d}{dt}\boldsymbol{x}(t)-\boldsymbol{f}_m(\boldsymbol{x},\boldsymbol{u};{\boldsymbol{\mu}}_1)$, resulting in the following regression problem:
\begin{equation}
    \label{sceq7}
        \delta\dot{\boldsymbol{X}}:=\dot{\boldsymbol{X}} - \boldsymbol{f}_m(\boldsymbol{X},\boldsymbol{U};{\boldsymbol{\mu}}_1)
    =\boldsymbol{\Theta}(\boldsymbol{X},\boldsymbol{U}) \boldsymbol{\Xi}.
\end{equation}
Note that the library $\boldsymbol{\Theta}(\boldsymbol{X},\boldsymbol{U})$ has the form
\begin{equation}
    \label{sceq8}
    \boldsymbol{\Theta}(\boldsymbol{X},\boldsymbol{U})=[{\theta}_1(\boldsymbol{X},\boldsymbol{U})\  {\theta}_2(\boldsymbol{X},\boldsymbol{U})\ \cdots {\theta}_v(\boldsymbol{X},\boldsymbol{U})],
\end{equation}
where $\theta_i(\boldsymbol{X},\boldsymbol{U})\in \mathbb{R}^{m\times 1}$ is a candidate function that may explain the discrepancy $\delta\boldsymbol{\phi}(t)$. 
The functions $\theta_i(\boldsymbol{X},\boldsymbol{U})$ can be any combination of $\boldsymbol{X}$ and $\boldsymbol{U}$. For example, $\theta_i(\boldsymbol{X},\boldsymbol{U})=\sin(\boldsymbol{X})\cos(\boldsymbol{U})$, $\theta_i(\boldsymbol{X},\boldsymbol{U})=\boldsymbol{X}\boldsymbol{U}^2$. By solving Eq.~\eqref{sceq5} we are able to model $\boldsymbol{g}(\boldsymbol{x}(t),\boldsymbol{u}(t);\delta{\boldsymbol{\mu}}_2)$.

\section{Examples}
\label{sec4}
In this section, we demonstrate applications of the proposed approach to discover model discrepancies. We start with an illustrative example and then apply this method to experimental data from a double pendulum on a cart.

\vspace{-1ex}
\subsection{Van der Pol oscillator}
To begin, we will focus on an illustrative example, the Van der Pol oscillator, and show how the SINDy method can be used to compensate  for both  parameter errors and model inadequacy. The Van der Pol oscillator is given by  
\begin{subequations}
\label{eq19}
\begin{align}
\dot{x}_1 &=x_2,\\
\dot{x}_2 &=\alpha\left(1-x_1^{2}\right) x_2-x_1,
\end{align}
\end{subequations}
with parameter $\alpha=0.5$. 

\subsubsection{Parameter mismatch}
\label{ParameterMis}
Suppose we do not know the true parameter $\alpha$, but instead have an approximation $\alpha_1=0.1$. 
It is possible to compensate for this model discrepancy caused by parameter mismatch.
We first gather the measurement data of the actual system, in this case by integrating the true dynamics using a fourth order Runge Kutta scheme. 

We integrate for $25$ time units with time step $\Delta t=0.01$ and initial condition is $\boldsymbol{x}_0=(0.5,0)$. Random Gaussian measurement noise with amplitude $0.01$ is added to the data.

We evaluate the model dynamics $\boldsymbol{f}(\boldsymbol{x}(t);\alpha_1)$, with the measured trajectory $\boldsymbol{x}(t)$ and the inaccurate parameter $\alpha_1$.  The discrepancy between the true system output and the model  is then given by
$\delta\dot{\boldsymbol{\boldsymbol{x}}}(t):=
\frac{d}{dt}{\boldsymbol{x}}(t)-\boldsymbol{f}(\boldsymbol{x}(t);\alpha_1)$.
The augmented error matrix is formed as

\begin{equation}
        \delta\dot{\boldsymbol{\boldsymbol{X}}}=
        \begin{bmatrix}
            \delta\dot{\boldsymbol{\boldsymbol{x}}}(t_0) & \delta\dot{\boldsymbol{\boldsymbol{x}}}(t_1) & \delta\dot{\boldsymbol{\boldsymbol{x}}}(t_2) & \hdots & \delta\dot{\boldsymbol{\boldsymbol{x}}}(t_m)
        \end{bmatrix}^T,
    \label{eq22}
\end{equation}
and the augmented state is 
\begin{equation}
        {\boldsymbol{\boldsymbol{X}}}=
        \begin{bmatrix}
            {\boldsymbol{\boldsymbol{x}}}(t_0) & {\boldsymbol{\boldsymbol{x}}}(t_1) & {\boldsymbol{\boldsymbol{x}}}(t_2) & \hdots & {\boldsymbol{\boldsymbol{x}}}(t_m)
        \end{bmatrix}^T.
    \label{eq23}
\end{equation}
The next step is to evaluate the library on the data. Here, we construct the library as 
\begin{equation}
\newcommand{\mymatrix}[1]{\ensuremath{\left\downarrow\vphantom{#1}\right.   
\overset{\xrightarrow[\hphantom{#1}]{\text{Functions}}}{#1}}}
        \boldsymbol{\Theta}(x_1,x_2)=
        \rotatebox{90}{\text{\tiny Time}}\mymatrix{\setlength{\arraycolsep}{2.5pt}
        \begin{bmatrix}
            | & | & | &  | &  |  &  |  &  |     \\
            1 & x_1 & x_2 & x_1x_2 & x_1^2x_2 & x_1x_2^2 &x_1^2x_2^2\\
            | & | & | &  | &  |  &  |  &  |
        \end{bmatrix}}\textcolor{red}{,}
    \label{eq24}
\end{equation}
with polynomial terms up to second order.
Finally, we form the sparse regression problem
\begin{equation}
    \delta\dot{\boldsymbol{X}}=\boldsymbol{\Theta}(x_1, x_2) \boldsymbol{\Xi},
    \label{eq25}
\end{equation}
and solve for the coefficient matrix $\boldsymbol{\Xi}$. Results are shown in the top row of Fig.~\ref{fig2}.
SINDy successfully identifies the model discrepancy. 

\subsubsection{Model Inadequacy (or structure mismatch)}
We now assume that the model discrepancy is caused by model inadequacy, where the model $\boldsymbol{f}_m$ is missing the linear term in the second equation of the Van der Pol system:
\begin{subequations}\label{eq26}
    \begin{align}
        \dot{x}_1 &=x_2,\\
        \dot{x}_2 &=\alpha\left(1-{x}_{1}^{2}\right) x_2.
    \end{align}
\end{subequations}
We assume that the parameter $\alpha=0.5$ is correct. 
The data, $\boldsymbol{x},\dot{\boldsymbol{x}},\delta\dot{\boldsymbol{x}}$ is collected and the library of functions is constructed as before.  
The sparse coefficient matrix $\boldsymbol{\Xi}$ is determined via SINDy, and the model discrepancy is successfully modeled as shown in the bottom row of Fig.~\ref{fig2}.

To summarize, the SINDy method successfully identifies the discrepancy between the underlying governing equation and an inaccurate model. The imperfect model is then combined with the SINDy discrepancy model, resulting in an accurate prediction of the true system dynamics. 

\vspace{-1ex}
\subsection{Experimental double pendulum on a cart}
We now demonstrate the use of SINDy to identify the time-dependent Hamiltonian function for the double pendulum on a cart from experimental data.
A non-dissipative pendulum is a quintessential example of a conservative system, which admits invariants such as the Hamiltonian function, from which the governing equations can be derived.
The conservative Hamiltonian is given by $H_c(\boldsymbol{q},{\boldsymbol{p}})=T + V$,
where $\boldsymbol{q}$ and ${\boldsymbol{p}}$ are the generalized position and momentum of the system, and $T$ and $V$ represent the kinetic and potential energy of the system; the Hamiltonian $H_c$ is the total energy, which is constant along a trajectory. 
Since we measure $\boldsymbol{q}$ and compute the time derivative $\dot{\boldsymbol{q}}$, we will represent $H_c$ as a function of  ${\boldsymbol{q}}$ and  $\dot{\boldsymbol{q}}$ in the following. 

Real-world mechanical systems are, however, generally not conservative, but instead exhibit friction and damping from joints and wind resistance. 
Thus, the total energy is not conserved, and instead decays over time without additional exogenous energy input. 
While the potential and kinetic energy terms can be easily formulated, dissipation terms can be more challenging to derive.
In this work, we seek to identify the time-dependent dissipative effects, by modeling the difference between the conservative model Hamiltonian and the measured energy of the system.  
In practice, the \emph{observed} energy is obtained by evaluating the idealized conservative Hamiltonian, consisting of the kinetic and potential terms $T$ and $V$, on the measured trajectory.  The \emph{model} energy is given by evaluating the idealized Hamiltonian on the initial condition.  Thus, the difference gives the energy dissipation, which we will model with SINDy:
\begin{equation}
\delta H(\boldsymbol{q}(t),\dot{\boldsymbol{q}}(t)) := 
H_m(\boldsymbol{q}(t),\dot{\boldsymbol{q}}(t)) - H_m(\boldsymbol{q}(0),\dot{\boldsymbol{q}}(0)).
    \label{eq28}
\end{equation}

\subsubsection{Problem Formulation}
\begin{figure}
\vspace{1ex}
\begin{center}
\includegraphics[width=\linewidth]{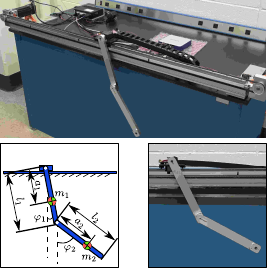}
\end{center}
\vspace{-2ex}
\caption{Double pendulum on a cart system with schematic (bottom left) and zoom of the two-link pendulum (bottom right). The pendulum cart is locked at its position to prevent horizontal movement.}
\label{DoublePenSystem}
\vspace{-4ex}
\end{figure}

We consider the double pendulum on a cart as shown in Fig.~\ref{DoublePenSystem}.
The kinetic and potential energy of the double pendulum (assuming a locked cart position) are
\begin{subequations}
\label{eq29a30}
\begin{align}
\begin{split}
    T&=\frac{1}{2}(m_1({\dot{x}_1}^2+{\dot{y}_1}^2)+m_2({\dot{x}_2}^2+{\dot{y}_2}^2))\\
    &\quad{}+\frac{1}{2}(I_1\dot{\varphi}_1^2+I_2\dot{\varphi}_2^2),
\end{split}
    \\
	V&=(m_1y_1+m_2y_2)g,
\end{align}
\end{subequations}
where $I_1$ and $I_2$ are the inertia, $m_1$ and $m_2$ are the masses, and $l_1$ and $l_2$ are the lengths of each pendulum arm, respectively.
$\varphi_1$ and $\varphi_2$ are the angles and $\dot{\varphi}_1$ and $\dot{\varphi}_2$ are the angular velocity of the first and second pendulum arm. The relative lengths to the center of mass of the first and second pendulum arm are given by $a_1$ and $a_2$.

The position of each pendulum arm's center of mass
$(x_1,y_1)$ and $(x_2,y_2)$ in~\eqref{eq29a30} can be determined as
\begin{subequations}
\label{eq31a34}
\begin{align}
x_1&=a_1\sin(\varphi_1),\\
x_2&=l_1\sin(\varphi_1)+a_2\sin(\varphi_2),\\
y_1&=a_1\cos(\varphi_1),\\
y_2&=l_1\cos(\varphi_1)+a_2\cos(\varphi_2).    
\end{align}
\end{subequations}

We assume that we can accurately determine the kinetic and potential energy of the system
\begin{equation}
    H_m(\varphi_1,\varphi_2,\dot{\varphi}_1,\dot{\varphi}_2)=T+V,
    \label{eq35}
\end{equation}
which represents the insufficient model for the total energy.
The discrepancy model $\delta H(\varphi_1,\varphi_2,\dot{\varphi}_1,\dot{\varphi}_2)$ then comprises the dissipative energy terms.

The frictional torque of the pendulum arm can be modeled as 
$\Gamma_{1}=k_{1}\dot{\varphi}_{1}$ and $\Gamma_{2}=k_{2}({\dot{\varphi}_{1}-\dot{\varphi}_{2}})$,

where $k_{1}$ and $k_{2}$ are damping coefficients. Then, $\delta H(\varphi_1,\varphi_2,\dot{\varphi}_1,\dot{\varphi}_2)$ is given by
\begin{equation}\label{eq38} 
\begin{split}
    \delta H(\varphi_1,\varphi_2,\dot{\varphi}_1,\dot{\varphi}_2) &=\int_{0}^{t}  \Gamma_{1}\dot{\varphi}_{1}+\Gamma_{2}({\dot{\varphi}_{1}-\dot{\varphi}_{2}}) dt \\
    &= \int_{0}^{t} k_{1}\dot{\varphi}_{1}^{2}+k_{2}({\dot{\varphi}_{1}-\dot{\varphi}_{2}})^{2} dt.
\end{split}     
\end{equation}

In many engineering applications, the direct measurement of the frictional term is difficult if not impossible. Thus, we would like to use our data-driven approach to learn a model for the dissipated energy~\eqref{eq38}.  Suppose that all the states $\varphi_1,\varphi_2,\dot{\varphi}_1,\dot{\varphi}_2$ can be measured or estimated, then $H_m(\varphi_1,\varphi_2,\dot{\varphi}_1,\dot{\varphi}_2)$ can be immediately calculated. 
Also, we can use the energy at the initial measurement time as reference for the total energy of the system, $H(\varphi_1(t_0),\varphi_2(t_0),\dot{\varphi}_1(t_0),\dot{\varphi}_2(t_0))=E_0$.

This allows us to calculate the dissipated energy as
\begin{equation}
\delta H(\varphi_1,\varphi_2,\dot{\varphi}_1,\dot{\varphi}_2)=E_0-H_m(\varphi_1,\varphi_2,\dot{\varphi}_1,\dot{\varphi}_2).
\label{eq39}
\end{equation}
To model this energy discrepancy, we define a library $\bTheta(\varphi_1,\varphi_2,\dot{\varphi}_1,\dot{\varphi}_2)$ that contains polynomial and Fourier terms up to the third order. Then, the discrepancy can be represented by
\begin{equation}
\delta H(\varphi_1,\varphi_2,\dot{\varphi}_1,\dot{\varphi}_2)=\bTheta(\varphi_1,\varphi_2,\dot{\varphi}_1,\dot{\varphi}_2)\boldsymbol{\xi},
\label{eq40}
\end{equation}
where $\boldsymbol{\xi}$ is a sparse vector that contains the coefficients of the active terms in the library.\\
\begin{table*}
\vspace{1ex}
\caption{The parameters of the experimental double pendulum system.}
\centering

\begin{tabular}{l|llllll}
\hline
{\color[HTML]{000000} Pendulum} & {\color[HTML]{000000} Mass ($kg$)} & {\color[HTML]{000000} Center of Mass ($m$)} & {\color[HTML]{000000} Inertia ($kg\cdot m^2$)} & {\color[HTML]{000000} Length ($m$)} & {\color[HTML]{000000} Damping Coefficient}              & Gravity Constant ($m/s^2$) \\ \hline
1st Arm                           & $0.2704     $                     & $0.1910         $                          & $0.003 $                                                   & $0.2667   $                        & $7.24\times{10^{-4}}$  & $9.818  $                                  \\ \hline
2nd Arm                          &$ 0.2056  $                        &$ 0.1621    $                               & $0.0011     $                                              & $0.2667  $                         & $1.65\times{10^{-4}}$ &                                          \\ \hline
\end{tabular}
\label{tab1}
\end{table*}
\begin{figure}
\vspace{-7ex}
	\centering
	\subfloat[Left: Initial energy of the system $E_0$ and the sum of kinetic and potential energy  $H_m(\varphi_1,\varphi_2,\dot{\varphi}_1,\dot{\varphi}_2)$, which decreases due to friction. Right: The difference between $E_0$ and $H_m(\varphi_1,\varphi_2,\dot{\varphi}_1,\dot{\varphi}_2)$ representing the energy dissipated by friction. ]{
		\label{fig4a}	
		\includegraphics[width=0.47\textwidth]{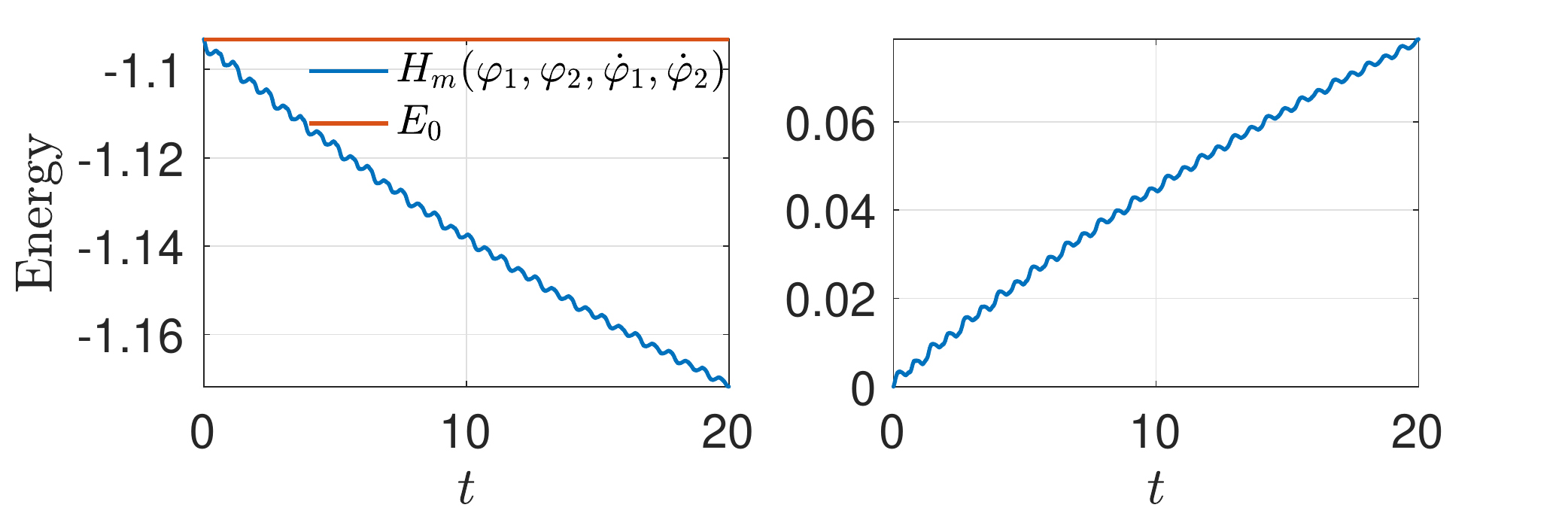}}\\
	\subfloat[The magnitude of energy dissipated by the friction $\delta H(\varphi_1,\varphi_2,\dot{\varphi}_1,\dot{\varphi}_2)$ and the SINDy identified dynamics.  Note that the discrepancy is positive, since it is defined as the ideal conserved energy minus the measured dissipative energy.   ]{
		\label{fig4b}	
		\includegraphics[width=0.47\textwidth]{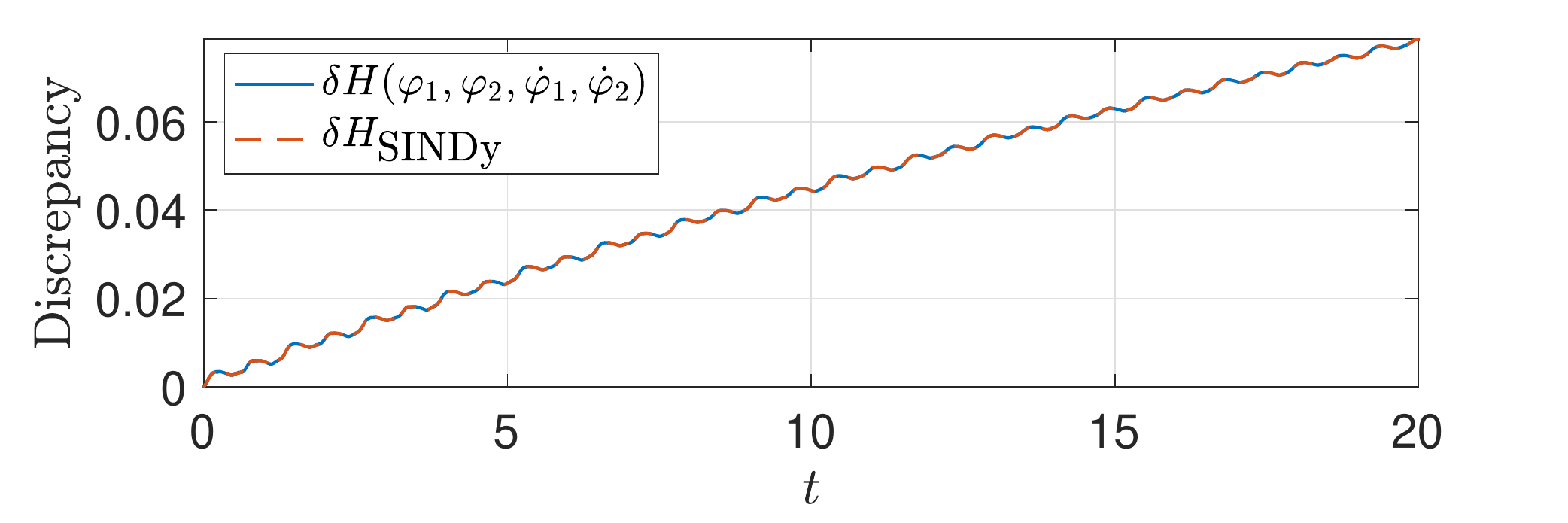}}\\ 
	\subfloat[The SINDy estimation error on the training data. The percentage is calculated using $\frac{\delta H(\varphi_1,\varphi_2,\dot{\varphi}_1,\dot{\varphi}_2)-\delta H_\text{SINDy}}{\max(| \delta H(\varphi_1,\varphi_2,\dot{\varphi}_1,\dot{\varphi}_2) |)}\times 100\%.$]{
		\label{fig4c}
		\includegraphics[width=0.47\textwidth]{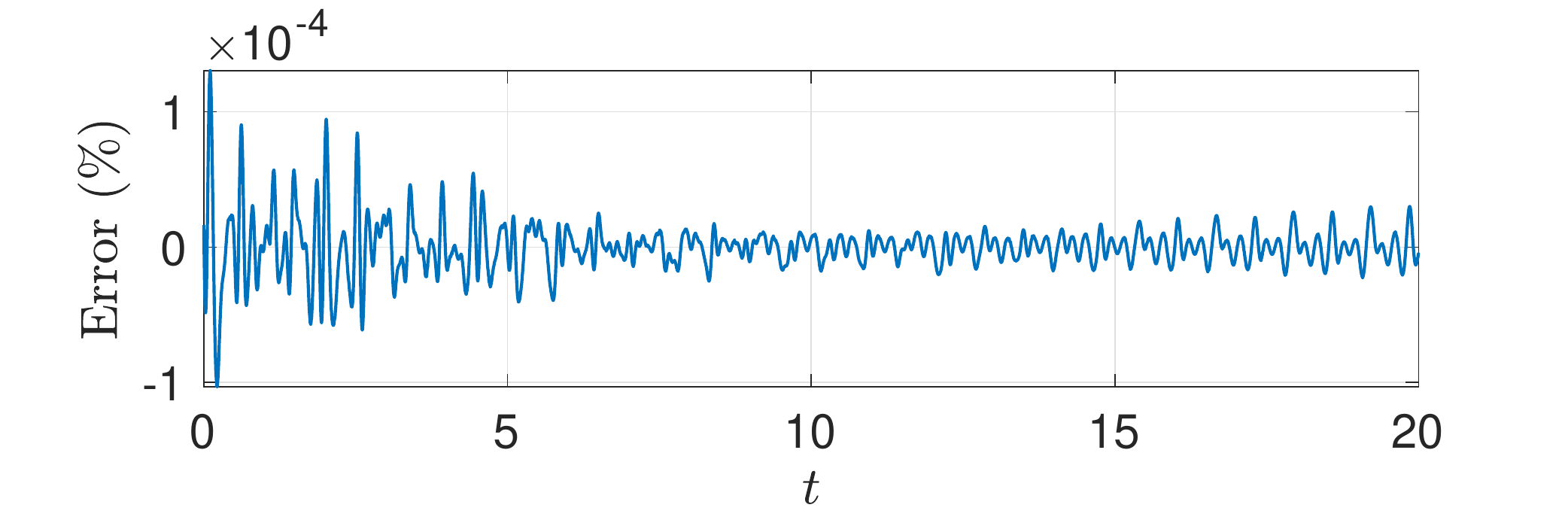}}\\
		\vspace{-1ex}
	\caption{
		Results for the identified model discrepancy caused by friction, demonstrated on training data.}
	\label{fig4}
	\vspace{-2ex}
\end{figure}
\begin{figure}
\vspace{-1ex}
    \centering
	\subfloat[The magnitude of energy dissipated by the friction $\delta H(\varphi_1,\varphi_2,\dot{\varphi}_1,\dot{\varphi}_2)$ and the SINDy estimation evaluated on validation data.]{
	\label{fig5a}
	\includegraphics[width=0.47\textwidth]{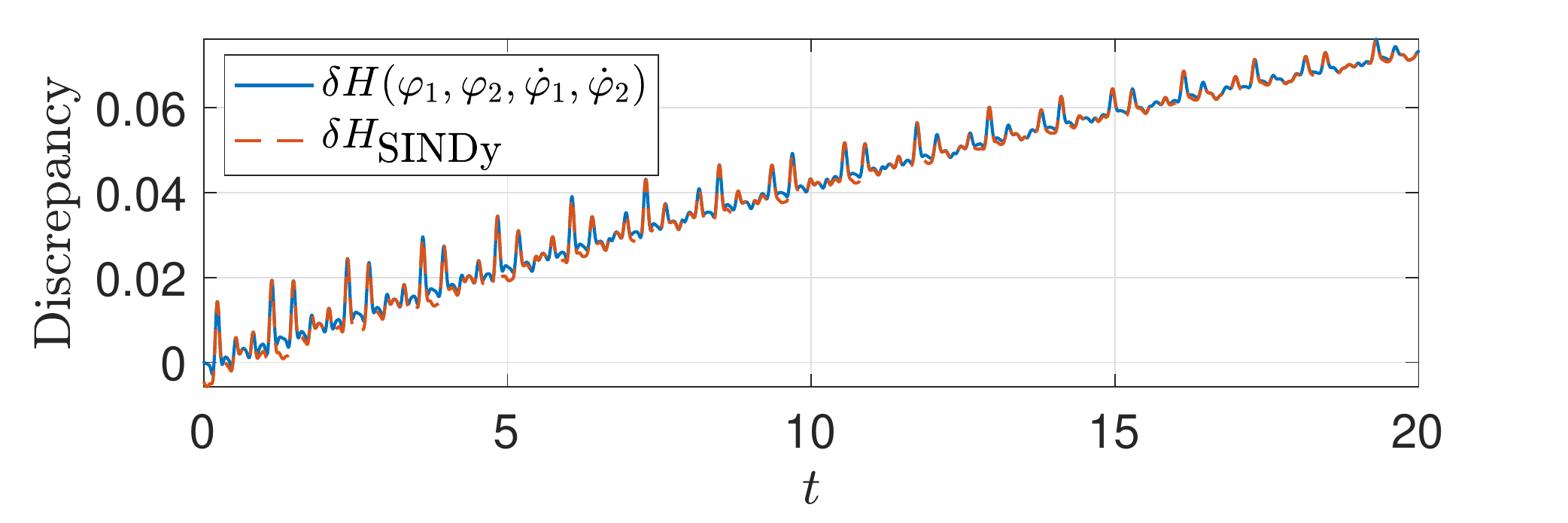}}\\
    \subfloat[The SINDy estimation error on validation data. The percentage is calculated using $\frac{\delta H(\varphi_1,\varphi_2,\dot{\varphi}_1,\dot{\varphi}_2)-\delta H_\text{SINDy}}{\max( |\delta H(\varphi_1,\varphi_2,\dot{\varphi}_1,\dot{\varphi}_2)| )}\times 100\%.$]{
	\label{fig5b}
	\includegraphics[width=0.47\textwidth]{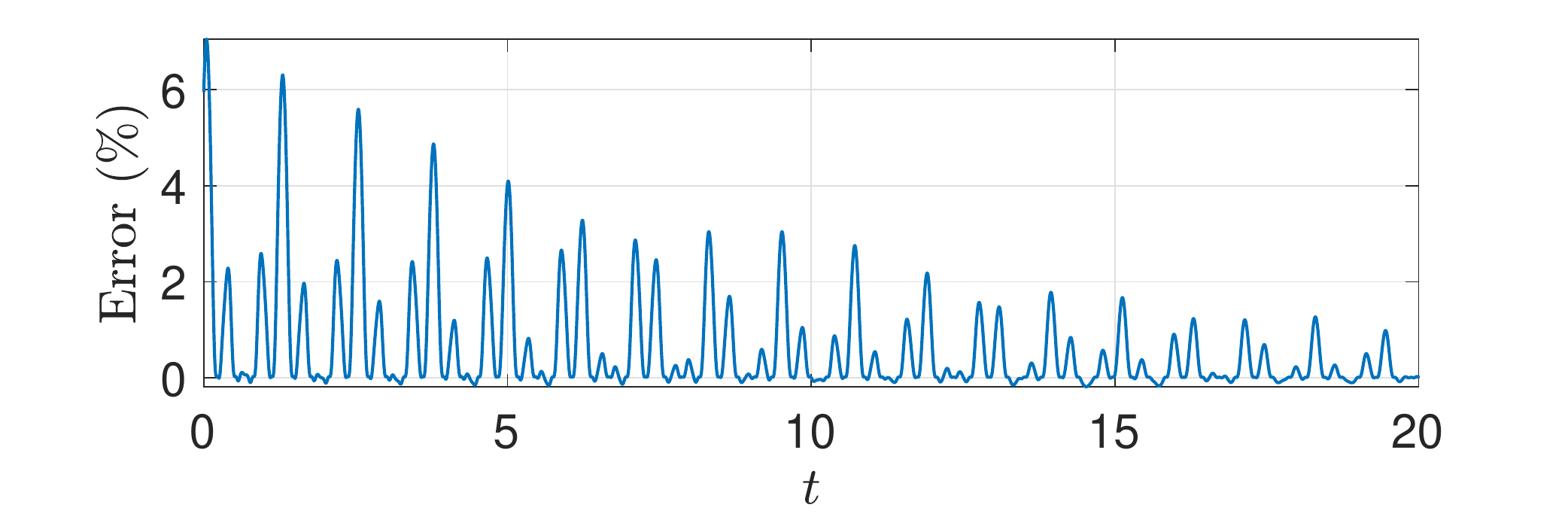}}\\
	\vspace{-1ex}
    \caption{
    Results for the identified model discrepancy caused by friction, demonstrated on validation data.}
    \label{fig5}
    \vspace{-4ex}
\end{figure}
\subsubsection{Results}
Here, we present results for modeling the dissipation 
$\delta H(\varphi_1,\varphi_2,\dot{\varphi}_1,\dot{\varphi}_2)$. The parameters of our experimental system are displayed in Table.~\ref{tab1}.  All parameters are obtained using a parameter estimation technique~\cite{DoublePenWingUp}. The pendulum mass and length are constrained during the parameter estimation so that there won't be a large deviation from the measured value.
The experiment was initialized at a random position, and the angles $\varphi_1$ and $\varphi_2-\varphi_1$ were collected with a sampling rate of $\Delta t=0.001$ for a duration of $20$s. 
We used a US Digital HUBDISK-1 1'' transmissive rotary encoder disk, which provides 5000 counts per revolution. 
Then the angular velocities $\dot{\varphi}_1$ and $\dot{\varphi}_2$ are approximated by taking numerical derivatives on the raw data.  
To mitigate noise, we first smooth the raw $\varphi_1$ and $\varphi_2$ data using the Savitzky-Golay filter and afterwards compute the derivative by numerical differentiation.

Results of the discrepancy model, identified with SINDy on the time series of  $\delta H(\varphi_1,\varphi_2,\dot{\varphi}_1,\dot{\varphi}_2)$, are shown in Fig.~\ref{fig4} for the training data and in Fig.~\ref{fig5} for validation data. Note that in Fig.~\ref{fig4a} the signal $H_{m}$ does not decrease asymptoticly. 
We observe that the table on which the pendulum in mounted oscillates with the pendulum, storing a small amount of potential energy.  The error of the SINDy prediction and the measured $\delta H(\varphi_1,\varphi_2,\dot{\varphi}_1,\dot{\varphi}_2)$ is shown in Fig.~\ref{fig4b}. From Fig.~\ref{fig4c} we see that SINDy accurately describes the effect of friction, explaining the model discrepancy.  

Crossvalidation is critical to avoid overfitting.
Hence, we test the identified model on a new validation dataset unseen during the training stage. The performance of the SINDy-discovered model on the validation data is shown in Fig.~\ref{fig5a}.  
Although the errors are larger on the validation dataset, the SINDy model is quite accurate in modeling the missing friction term, demonstrating the ability of the discrepancy model to generalize to new test cases.

\subsection{Double pendulum on a cart with control}
\begin{figure*}
\vspace{1ex}
    \centering
    \includegraphics[width=0.999\textwidth]{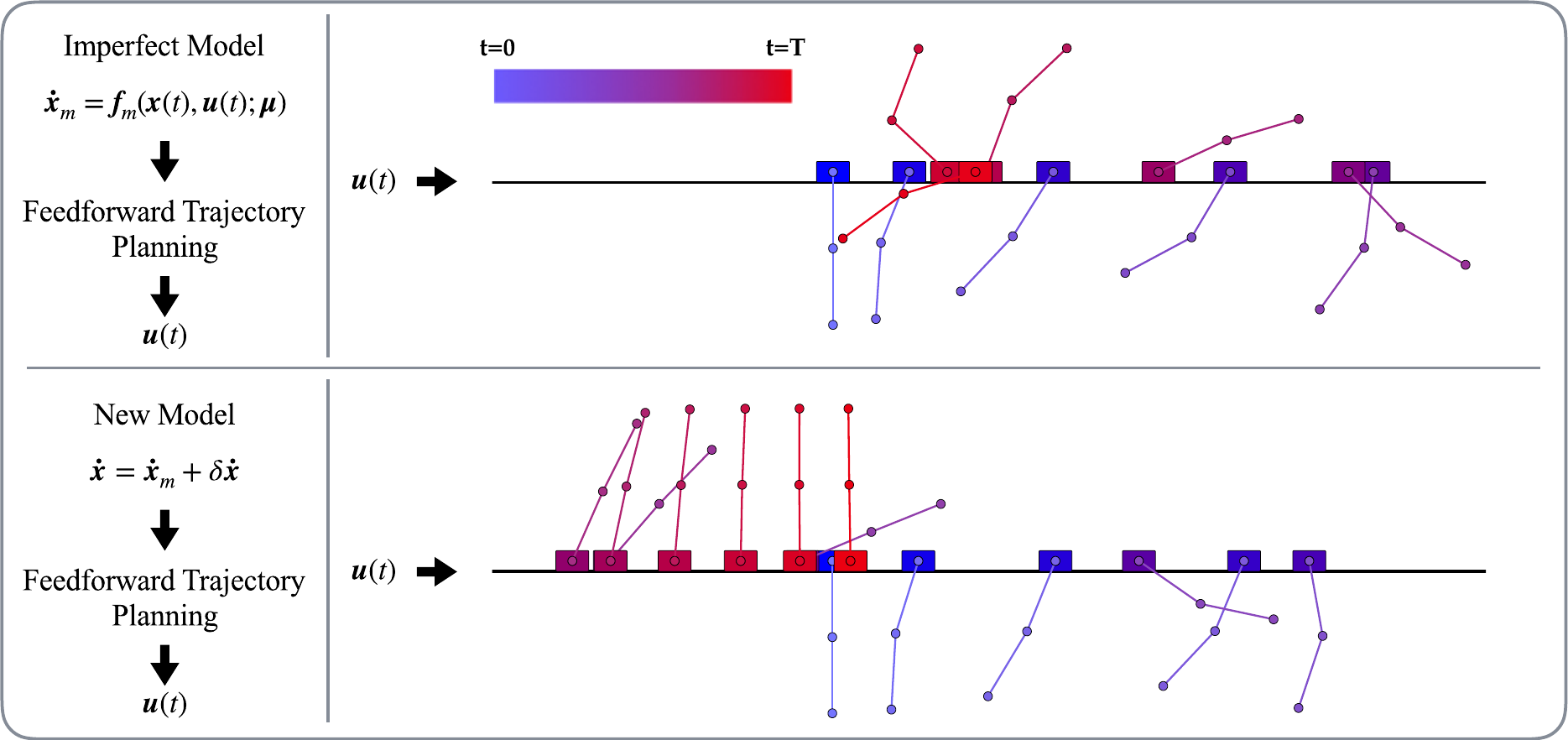}
    \caption{Swing-up of the double pendulum on a cart using feed-forward control. 
    Top: The feed-forward control input is calculated using the imperfect model, which is in then applied to the actual system. The pendulum cannot reach the up-right position. 
    Bottom: System's response under the control input calculated using the hybrid model, which is comprised of the imperfect model and the identified discrepancy model. Since the discrepancy is identified, the new model mimics the true dynamics of the double pendulum on a cart system.}
    \label{PenUp}
    \vspace{-4ex}
\end{figure*}

In the previous section, SINDy is utilized to discover discrepancy models in the total energy caused by dissipation when given an insufficient conservative energy model. 
We now consider the case of an actuated double pendulum on a cart, where a control input is added to the system.  
This is an important generalization, since many real-life system are affected by control, which requires additional modeling. To discover the model discrepancy in systems with control, we follow the steps outlined in Sec.~\ref{SecSINDyc}.

\subsubsection{Problem formulation}
In this example, we consider data from a numerical simulation of the double pendulum on a cart, shown in Fig.~\ref{DoublePenSystem}, with parameters provided in Tab.~\ref{tab1}. The equations of motion can be represented as 
\begin{equation}
    \dot{\boldsymbol{x}}= \boldsymbol{f}(\boldsymbol{x})+ [0,0,0,0,0,1]^T u,
\end{equation}

where
$\boldsymbol{x}=[\varphi_1,\varphi_2,s,\dot{\varphi_1},\dot{\varphi_2},\dot{s}]^T$ is the state vector and $s$ represents the displacement of the pendulum cart. 
We choose the acceleration of the pendulum cart as control input so that
$u=\ddot{s}$. 
We refer to~\cite{DoublePenWingUp} for a derivation of the equations of motion and a technique to estimate parameters from experimental data. 

We seek to perform the swing-up control of the double pendulum. 
However, the employed model is flawed, as it contains an additional term and an incorrect parameter:
\begin{equation}
{\small\hspace{-0.15cm}\boldsymbol{f}_m(\boldsymbol{x}) =\hspace{-0.1cm} \boldsymbol{f}(\boldsymbol{x}) 
\hspace{-0.1cm}+\hspace{-0.1cm} [\sin(\varphi_1),0,0,0,0,0]^T 
\hspace{-0.1cm}+\hspace{-0.1cm} [0,0,0,0,0,0.95]^T u}.
\end{equation}

The discrepancy model is then
\begin{equation}
    \delta \dot{\boldsymbol{x}}=\dot{\boldsymbol{x}}-\boldsymbol{f}_m(\boldsymbol{x})=[-\sin(\varphi_1),0,0,0,0,0.05u]^T, 
    \label{eq45}
\end{equation}
which we seek to model.

First, we generate data for the identification using the imperfect model.  
Specifically, we design a feed-forward control input to swing up the double pendulum based on the imperfect model $\boldsymbol{f}_m(\boldsymbol{x})$.  
When this control signal is applied to the actual system, the observed behavior deviates from the pre-planned trajectory due to the model discrepancy. 
This difference is then used to form a sparse regression problem, as in Eq.~\eqref{sceq8}, to identify the discrepancy $\delta \dot{\boldsymbol{{x}}}$. 
Finally, a new swing-up trajectory is designed using the hybrid model consisting of the imperfect model, augmented with the discrepancy model.

\subsubsection{Results}
Simulation results for the swing-up of the double pendulum are shown in Fig.~\ref{PenUp}. 
The feed-forward trajectory is determined as the solution to an optimization problem~\cite{ParnMPC}. The simulation time step is $\Delta t=0.001$, the prediction horizon is $2$, and the total swing-up time is chosen to be $T=2.7$. The weight matrices are $Q=\text{diag}[10,10,20,1,1,0.1]$ and $R=1$ for the state and input, respectively. It is demonstrated in Fig.~\ref{PenUp} that SINDy correctly identifies the model mismatch shown in Eq.~\eqref{eq45} and that the new model mimics the actual dynamics of the system, which results in a successful swing-up of the pendulum.

\section{Conclusion}
\label{sec5}
In this work, we present a data-driven framework to model discrepancies between observations and simplified or incorrect physical models.  In particular, we leverage the sparse identification of nonlinear dynamics (SINDy)~\cite{SINDY} algorithm to discover model discrepancies caused by parameter mismatch or model inadequacy. 
We demonstrate this approach on several systems, including the Van der Pol oscillator and experimental and numerical data from a double pendulum on an actuated cart.

Our results suggest that a hybrid discrepancy modeling approach, involving a physics-based model and a data-driven correction, may have several benefits.  
First, we incorporate prior knowledge and enforce conservation laws and constraints that are notoriously challenging in data-driven approaches. In addition, focusing SINDy on the model mismatch is a much simpler task than trying to model all system dynamics at once, which would involve a large library of candidate terms and a potentially ill-conditioned inverse problem. 
This ill-conditioned problem often leads to the mis-estimation of parameters, such as the mass and length of the pendulum arms, which are accurately measured ahead of time.   
Instead, we are able to focus the data-driven effort on modeling the few terms and parameters that cause the discrepancy. 

There are several future directions suggested by this work.  
First, it will be important to extend this framework to include noise and exogenous disturbances, as in the Kalman filter, along with partial measurements.  
It will also be interesting to use our discrepancy models for the experimental swing-up control of the double pendulum on a cart, which is the subject of ongoing work.

\section*{Acknowledgment}
SLB would like to acknowledge funding support from the Army Research Office (ARO W911NF-17-1-0306, W911NF-17-1-0422, and W911NF-19-01-0045). EK gratefully acknowledges support by the ``Washington Research Foundation Fund for Innovation in Data-Intensive Discovery" and a Data Science Environments project award from the Gordon and Betty Moore Foundation (Award \#2013-10-29) and the Alfred P. Sloan Foundation (Award \#3835) to the University of Washington eScience Institute, and funding through the Mistletoe Foundation.
JNK and SLB acknowledge support from the Defense Advanced Research Projects Agency (DARPA- PA-18-01-FP-125).
We would  like to acknowledge valuable discussions with Brian DeSilva, Tony Piaskowy, and Aditya Nair.

\bibliographystyle{plain}        
\bibliography{PaperReference} 

\end{document}